\documentclass{article}

\usepackage{microtype}
\usepackage{graphicx}
\usepackage{subfigure}
\usepackage{booktabs} % for professional tables

\usepackage{hyperref}
\usepackage{multirow}

\newcommand{\numstd}[2]{
    #1{\scriptsize$\pm$#2}
}
\usepackage[accepted]{icml2025}

\usepackage{amsmath}
\usepackage{amssymb}
\usepackage{mathtools}
\usepackage{amsthm}
\usepackage{makecell}

\usepackage[capitalize,noabbrev]{cleveref}

\theoremstyle{plain}
\newtheorem{theorem}{Theorem}[section]

\theoremstyle{definition}

\theoremstyle{remark}

\usepackage[textsize=tiny]{todonotes}

\icmltitlerunning{Kolmogorov-Arnold Networks for Time Series Granger Causality Inference}

\begin{document}

\twocolumn[
\icmltitle{Kolmogorov-Arnold Networks for Time Series Granger Causality Inference }



\icmlsetsymbol{equal}{*}

\begin{icmlauthorlist}
\icmlauthor{Meiliang Liu}{yyy}
\icmlauthor{Yunfang Xu}{yyy}
\icmlauthor{Zijin Li}{yyy}
\icmlauthor{Zhengye Si}{yyy}
\icmlauthor{Xiaoxiao Yang}{yyy}
\icmlauthor{Xinyue Yang}{yyy}
\icmlauthor{Zhiwen Zhao}{yyy}
\end{icmlauthorlist}

\icmlaffiliation{yyy}{School of Artificial Intelligence, Beijing Normal University, China}


\icmlkeywords{Machine Learning, ICML}

\vskip 0.3in
]




\begin{abstract}
We propose the Granger causality inference Kolmogorov-Arnold Networks (KANGCI), a novel architecture that extends the recently proposed Kolmogorov-Arnold Networks (KAN) to the domain of causal inference. By extracting base weights from KAN layers and incorporating the sparsity-inducing penalty and ridge regularization, KANGCI effectively infers the Granger causality from time series. Additionally, we propose an algorithm based on time-reversed Granger causality that automatically selects causal relationships with better inference performance from the original or time-reversed time series or integrates the results to mitigate spurious connectivities. Comprehensive experiments conducted on Lorenz-96, Gene regulatory networks, fMRI BOLD signals, VAR, and real-world EEG datasets demonstrate that the proposed model achieves competitive performance to state-of-the-art methods in inferring Granger causality from nonlinear, high-dimensional, and limited-sample time series.
\end{abstract}

\section{Introduction}
\label{submission}
Granger causality is a statistical framework for analyzing the causal relationship between time series. It offers a powerful tool to investigate temporal dependencies and the direction of influence between variables \cite{11seth2007granger,12maziarz2015review,13friston2014granger,14shojaie2022granger}. By examining the past values of time series, Granger causality seeks to determine if the historical knowledge of one variable improves the prediction of another \citep{15bressler2011wiener,16barnett2014mvgc}. Revealing inner interactions from time series has made Granger causality useful for the investigation in many fields, such as econometrics \citep{17mele2022innovation}, neuroscience \citep{18chen2023granger}, climate science \citep{19ren2023impact}, etc.  

Recently, there has been a growing interest in incorporating the neural network into the study of Granger causality due to its inherent nonlinear mapping capabilities. For now, a variety of neural Granger causality models have been proposed, mainly based on multi-layer perceptron (MLP) \cite{1tank2021neural,8bussmann2021neural,6zhoujacobian}, recurrent neural network (RNN) \cite{3khanna2019economy,1tank2021neural}, convolutional neural network (CNN) \cite{2nauta2019causal}, or their combination \cite{5cheng2024cuts+}. These models have achieved significant improvements in inferring nonlinear Granger causality but still have some limitations:  (1) RNN-based models are more suitable for processing long time series but experience decreased inference performance in the limited time-sample scenario. (2) MLP-based models face the challenge of low inference efficiency when dealing with high-dimensional and noisy time series. (3) CNN-based models perform ineffectively on many nonlinear datasets.

Therefore, our motivation is to propose a neural network-based Granger causality model that can effectively infer causal relationships from high-dimensional nonlinear time series with limited sampling points. We consider a novel framework, the Kolmogorov-Arnold Network (KAN) \cite{7liu2024kan}, to construct a Granger causality inference model. Different from MLP, which uses learnable weights on the edges and fixed activation functions on the nodes, KAN uses learnable univariate functions at the edges and simple summation operations at the nodes, making its computational graph much smaller than that of MLP \cite{20kiamari2024gkan,21hou2024comprehensive}. 

Our work extends the basic KAN to the field of causal inference and aims to evaluate whether the KAN-based model has the potential to outperform MLP-based and RNN-based baselines. Our main contributions are as follows:

\begin{itemize}
  
    \item We propose a simple but effective Granger causality model based on KAN. The model only needs to extract base weights of KAN layers and impose the sparsity-inducing penalty and ridge regularization to infer Granger causality.

    \item We propose an algorithm that automatically selects the Granger causality adjacency matrix with the higher inference performance from the origin or time-reversed time series or mitigates spurious connections by fusing both of them. 

    \item Extensive experiments on Lorenz-96, Gene regulatory networks, fMRI BOLD, VAR, and real-world EEG datasets validate that the proposed model attains stable and competitive performances in Granger causality inference.
\end{itemize}

\section{Background and Related Works}
\label{Related Works}

\subsection{Background: Neural network-based Granger causality}
Inferring Granger causality from nonlinear time series via neural networks has attracted widespread attention. \citet{1tank2021neural} proposed the cMLP and cLSTM, which extracted the first-layer weights of MLP and long short-term memory (LSTM) and imposed the sparsity-inducing penalty to infer Granger causality. \citet{8bussmann2021neural} proposed the Neural Additive Vector Autoregression (NAVAR) model based on MLP and LSTM, called NAVAR(MLP) and NAVAR(LSTM), for Granger causality inference. \citet{3khanna2019economy} proposed the economy-SRU (eSRU) model, which extracted weights from statistical recurrent units (SRU) and also imposed sparsity-inducing penalty to infer Granger causality. \citet{2nauta2019causal} proposed the Temporal Causal Discovery Framework (TCDF) based on temporal convolutional network (TCN) and causal verification algorithm to infer Granger causality and select time lags. \citet{9cheng2023cuts} proposed Causal discovery from irregUlar Time-Series (CUTS), which could effectively infer Granger causality from time series with random missing or non-uniform sampling frequency. Subsequently, to solve the problems of large causal graphs and redundant data prediction modules of CUTS, \citet{5cheng2024cuts+} proposed the CUTS+, which introduced a coarse-to-fine causal discovery mechanism and a message-passing graph neural network (MPGNN) to achieve more accurate causal reasoning. \citet{4marcinkevivcs2021interpretable} proposed the generalised vector autoregression (GVAR) based on the self-explaining neural network model, which effectively inferred causal relationships and improved the interpretability of the model. \citet{6zhoujacobian} proposed a neural Granger causality model based on Jacobi regularization (JRNGC), which only needs to construct a single model for all variables to achieve causal inference.

\subsection{Related Works}
\subsubsection{Component-wise nonlinear autoregressive (NAR)}
Assume a $p$-dimensional nonlinear time series $x_{t}=[x_{<t1},\ldots ,x_{<tp}]$, where $x_{<ti}=(\ldots,x_{<(t-2)i},x_{<(t-1)i})$. In the nonlinear autoregressive (NAR) model, the ${t^{th}_{}}$ time point $x_{t}$ can be denoted as a function $g$ of its past time values:
\begin{equation}
\label{equation3}
x_{t}=g\left( x_{<t1},\ldots ,x_{<tp}\right)  +e^{t}
\end{equation}

Furthermore, in the component-wise NAR model, it is assumed that the ${t^{th}_{}}$ time point of each time series $x_{ti}$ may depend on different past-time lags from all the series:
\begin{equation}
\label{equation5}
    x_{ti}=g_{i}\left( x_{<t1},\ldots ,x_{<tp}\right)  +e^{ti}
\end{equation}
To infer Granger causality from the component-wise NAR model, sparsity-inducing penalty is applied:
\begin{equation}
\label{equation6}
\begin{split}
        \min_{W} \sum^{T}_{t=K} \left( x_{ti}-g_{i}\left( x_{<t1},\ldots ,x_{<tp}\right)  \right)^{2} \\ +\lambda \sum^{p}_{j=1} \Theta \left( W_{:,j}\right)  
\end{split}
\end{equation}
where ${W}$ is extracted from the neural network, $\Theta$ is the sparsity-inducing penalty that penalizes the parameters in ${W}$ to zero, $\lambda$ is the hyperparameter that controls the strength of the penalty. In the NAR model, if there exists a time lag $k$, ${W^{k}_{:,j}}$ contains non-zero parameters, time series $j$ Granger-causes to time series $i$.

\subsubsection{Time reversed Granger causality}
The time-reversed Granger causality is initially introduced by \citet{23haufe2013critical}, which is used to reduce spurious connections caused by volume conduction effects in analyzing Electroencephalogram (EEG) signals \cite{24van1998volume,25nunez1997eeg}. Subsequently, \citet{26winkler2016validity} demonstrate that, in finite-order autoregressive processes, causal relationships would reversed in time-reversed time series. Moreover, comparing the causal relationship inferred from the original and time-reversed time series can enhance the robustness of causal inference against noise. However, the findings of \citet{26winkler2016validity} primarily apply to linear systems. Recent research indicates that in nonlinear chaotic systems, causal relationships inferred from time-reversed time series generally align with those from the original data, with perfect causal relationship reversal occurring only under specific conditions \cite{10kovrenek2021causality}.

\subsubsection{Kolmogorov–Arnold Networks (KAN)}
\citet{7liu2024kan} proposed KAN, which has garnered attention as a compelling alternative to MLP. The theoretical foundation of MLP is rooted in the universal approximation theorem, which demonstrates that neural networks can approximate any continuous function under appropriate conditions \cite{27pinkus1999approximation}. By contrast, KAN is grounded in the Kolmogorov-Arnold (KA) representation theorem,  which states that any multivariate continuous function can be represented by the sum of a finite number of univariate functions \cite{28schmidt2021kolmogorov}.
\begin{theorem}
\label{thm_KAN}
Let \( f:[0,1]^n\to \mathbb{R}\) be a continuous multivariate function.  
There exist continuous univariate functions \(\Phi_q\) and \( \phi_{q,p} \) such that:  
\[f(x_1, x_2,\dots, x_n) =\sum_{i=1}^{2n+1} \Phi_q \left( \sum_{j=1}^{n} \phi_{q,p}(x_p) \right)\]
where \( \Phi_i:\mathbb{R}\to \mathbb{R}\) and \( \phi_{q,p}:[0, 1] \to \mathbb{R} \) are continuous functions.
\end{theorem}

Although the KA representation theorem is elegant and general, its application in deep learning remains limited before the work of \citet{7liu2024kan}. This limitation can be attributed to two primary factors: (1) the function $\phi_{q,p}$ is typically non-smooth; (2) the theorem is constrained to construct shallow neural networks with two-layer nonlinear architectures with limited hidden layer size. \citet{7liu2024kan} do not strictly constrain the neural network to fully adhere to Theorem \ref{thm_KAN}, but instead extend the network to arbitrary width and depth, making it applicable for deep learning. Due to this alternation, KAN and its variants have been extensively applied across various domains, including computer vision \cite{29bodner2024convolutional}, time series forecasting \cite{30xu2024kolmogorov}, health informatics \cite{31li2024u}. In this study, we develop our Granger causality model based on the code of efficientKAN \footnote{https://github.com/Blealtan/efficient-kan}.

\section{Model Architecture}

\subsection{Component-wise KAN}

\begin{figure*}[h]
\centering
\includegraphics[width=0.95\textwidth]{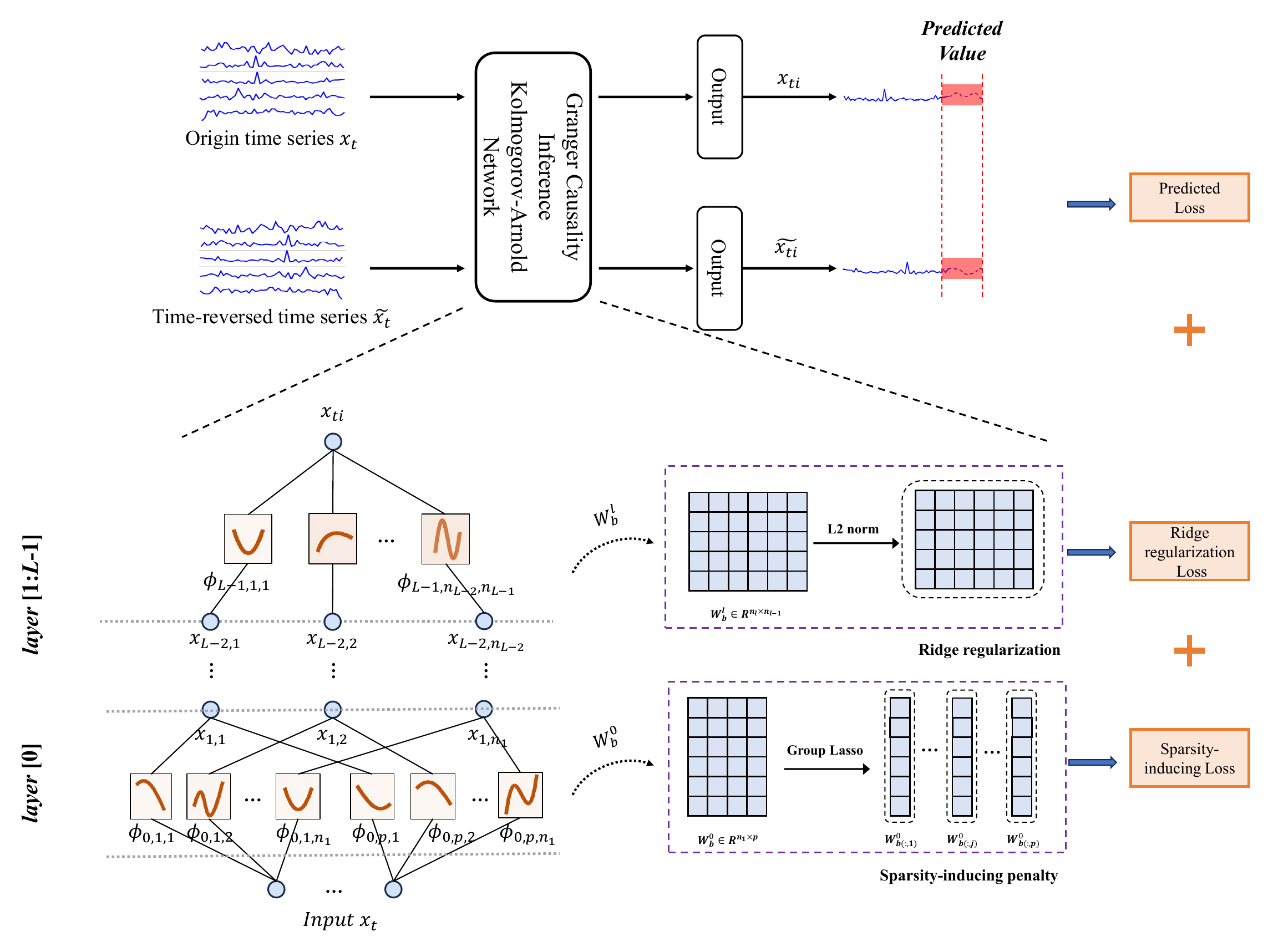}
\caption{The architecture of KANGCI.}
\label{IAF&ITF}
\end{figure*}

To extract the influence from input to output, we model each component $g_{i}$ using a separate KAN. Let $g_{i}$ take the form of a KAN with $L-1$ layers, and $h^{l}$ are denoted as the $l^{th}$ hidden layer. The trainable parameter of KAN including base weight $W_{base}$ and spline weight $W_{spline}$ on each layer. $W_{base}=\{W_{b}^{0},W_{b}^{1},\ldots,W_{b}^{L-1}\}$ and $W_{spline}=\{W_{s}^{0},W_{s}^{1},\ldots,W_{s}^{L-1}\}$. We separate the $W_{base}$ into the first layer weighs $W_{b}^{0}\in\mathbb{R}^{H\times p}$, and the other layers $W_{b}^{l}\in\mathbb{R}^{H\times H}$ ($0<l<L$). By using these notations, the vector of the hidden units in the first layer $h^{1}$ is denoted as:
\begin{equation}
\mathbf{h}^{1}=\underbrace{\left(\begin{array}{ccc}
\phi_{0, 1,1}(\cdot)           &\cdots    & \phi_{0, 1, n_{0}}(\cdot) \\
\phi_{0, 2,1}(\cdot)           & \cdots   & \phi_{0, 2, n_{0}}(\cdot) \\
\vdots                         & \vdots   &                            \\
\phi_{0, n_{1}, 1}(\cdot)    & \cdots   & \phi_{0, n_{1}, n_{0}}(\cdot)
\end{array}\right)}_{\boldsymbol{\Phi}_{0}} \mathbf{x}_{t}
\end{equation}

where $n_{0}=p$ is the input time series dimension, $n_{1}$ is the first hidden layer size. Here, the $\phi(x)$ is denoted as:
\begin{equation}
    \phi(x)=W_{b}^{0}b(x)+W_{s}^{0}spline(x)
\end{equation}
\begin{equation}
    b(x)=silu(x)=\frac{x}{1+e^{-x}}
\end{equation}
\begin{equation}
    spline(x)=\sum_{i}c_{i}B_i(x)
\end{equation}
where $b(x)=silu(x)$ is the Sigmoid Linear Unit activation function, $B_i$ is denoted as B-splines, $c_{i}$ is the control points (coefficients). Subsequently, The vector of the hidden units in the layer $l$ is denoted as:
\begin{equation}
\mathbf{h}^{l}=\underbrace{\left(\begin{array}{ccc}
\phi_{l-1, 1,1}(\cdot)           &\cdots    & \phi_{l-1, 1, n_{l-1}}(\cdot) \\
\phi_{l-1, 2,1}(\cdot)           & \cdots   & \phi_{l-1, 2, n_{l-1}}(\cdot) \\
\vdots                         & \vdots   &                            \\
\phi_{l-1, n_{l}, 1}(\cdot)    & \cdots   & \phi_{l-1, n_{l}, n_{l-1}}(\cdot)
\end{array}\right)}_{\boldsymbol{\Phi}_{l-1}} \mathbf{h}^{l-1}
\end{equation}
where $n_{l}$ and $n_{l-1}$ is the $l^{th}$ and ${l-1}^{th}$ hidden layer size, respectively. Here, the $\phi(x)$ is denoted as:
\begin{equation}
    \phi(x)=W_{b}^{l-1}b(x)+W_{s}^{l-1}spline(x)
\end{equation}
The time series $x_{t}$ go through the $L-1$ hidden layers to generate the output $x_{ti}$, which is denoted as:
\begin{equation}
\label{xti}
    x_{ti}=g_i(x_{t})+e_{ti}=\Phi_{L-1}\circ h^{L-1}+e_{ti}
\end{equation}
where $e_{ti}$ is the mean zero Gaussian noise.

\subsection{Applying sparsity-inducing penalty and ridge regularization on KAN to infer Granger causality}

According to Eq.\ref{equation6}, the inference of Granger causality in Eq.\ref{xti} uses component-wise NAR combined with sparsity-inducing penalty. In our study, we extract the base weight of the first hidden layer $W_{b}^{0}$ and apply the group lasso penalty to the columns of the $W_{b}^{0}$ matrices for each $g_{i}$, which is denoted as:
\begin{equation}
    GroupLasso(W_{b(:,j)}^0)=\left\|W_{b(:,j)}^0\right\|_{F}
\end{equation}
where $W_{b(:,j)}^0$ is the $j$ column of the $W_{b}^{0}$ corresponding to the time series $j$. $\|\cdot\|_{F}$ is denoted as the Frobenius matrix norm. The sparsity-inducing loss $\mathcal{L}_{s}$ is defined as:
\begin{equation}
    \mathcal{L}_{s}=\lambda\sum^{p}_{j=1}\| W^{0}_{b(:,j)}\|_{F} 
\end{equation}
$\lambda>0$ is the group lasso hyperparameter that controls the penalty strength. For the base weight of other hidden layers $W_{b}^l$, we apply ridge regularization to them, which is denoted as:
\begin{equation}
    RidgeRegularization(W_b^{1:L-1})=\sum_{l=1}^{L-1}\| W^{l}_{b}\|_{2} 
\end{equation}
where $\|\cdot\|_{2}$ is denoted as the $L2$ norm. The ridge regularization loss $\mathcal{L}_{r}$ is defined as:
\begin{equation}
    \mathcal{L}_{r}=\gamma\sum_{l=1}^{L-1}\| W^{l}_{b}\|_{2} 
\end{equation}
$\gamma>0$ is the ridge regularization hyperparameter that controls the regularization strength. Finally, the predicted loss is defined as:
\begin{equation}
        \mathcal{L}_{p}=\sum^{p}_{i=1} (x_{ti}-g_{i}(x_t))^{2}
\end{equation}
Therefore, the loss function is defined as:
\begin{equation}
    \mathcal{L}=\mathcal{L}_{p}+\mathcal{L}_{s}+\mathcal{L}_{r}
\end{equation}

Since the proposed model is a component-wise architecture, a total of $p$ models are needed to construct the complete Granger causality matrix. We extract the first hidden layer weight $W^{0}_b$ to compute the $i^{th}$ row of the Granger causality matrix $G$, which is denoted as:
\begin{equation}
\label{eq15}
    G_{(i,:)}=\|W^0_{b(:,j)}\|_{F}
\end{equation}

\subsection{Fusion of origin and time reversed time series}
During the experiment, we observe that, in certain simulation trials, the causal relationship inferred from the original and time-reversed time series exhibit considerable divergence. Specifically, there are cases where the performance inferred from the original time series is higher, while in other cases, the time-reversed time series yielded a better score. Consequently, our objective is to develop an algorithm that can automatically select the matrix with the higher causality score from either the original or reversed time series or obtain more accurate inference results by fusing both of them.

Algorithm \ref{alg1} summarizes the proposed algorithm for fusing the original and time-reversed time series. In the Granger causality inference stage, a total of 2$p$ KANGCI models are required, with the first $p$ models applied to the original time series and the next $p$ models to the time-reversed time series (lines 3-4 in Algorithm \ref{alg1}). Then, we use Eq.\ref{eq15} to calculate the Granger causality matrix for the original and reversed time series, respectively (line 5 in Algorithm \ref{alg1}). Subsequently, we compare the losses to determine whether to select a single matrix or fuse both matrices. Specifically, when the prediction loss and sparsity-inducing loss of the original time series are both lower than those of the reversed time series, it indicates that the model performs better in terms of prediction and sparsity on the original time series. Therefore, the Granger causality inferred from the original time series is chosen as the final result. Conversely, if the model exhibits lower losses on time-reversed time series, the Granger causality inferred from time-reversed time series is selected (lines 7-10 in Algorithm \ref{alg1}). In situations where the prediction loss and sparsity-inducing loss do not align between the two time series, we element-wise compare each element in the two matrices. If the absolute difference between the corresponding elements is below a predefined threshold (unified set to 0.05 in our study), the average of the two elements is taken. If the difference exceeds the threshold, the maximum value of the two elements is taken (lines 12-21 in Algorithm \ref{alg1}). In our experiments, this straightforward strategy can effectively improve the Granger causality inference performance of the proposed model.

\begin{algorithm}[!htb]
   \caption{Fusion of origin and time reversed time series for inferring Granger causality with KANGCI}
   \label{alg1}
\begin{algorithmic}[1]
   \STATE {\bfseries Input:} The origin multivariate time series $\{x_t\}$ with dimension $p$; group lasso penalty hyperparameter $\lambda$; ridge regularization hyperparameter $\gamma$; threshold $\theta$
   \STATE {\bfseries Output:} Estimate $\hat{G}$ of the adjacency matrix of the GC graph.   
   \STATE Let $\{\tilde{x_t}\}$ be the time-reversed time series of $\{x_t\}$, $\{x_1,x_2,\ldots,x_T\}\equiv\{\tilde{x_T},\tilde{x_{T-1}},\ldots,\tilde{x_1}\}$.   
   \STATE Train 2$p$ KANGCI with hyperparameter $\lambda$ and $\gamma$ (first $p$ models are trained on $\{x_t\}$ and next $p$ models are trained on $\{\tilde{x_t}\}$).  
   \STATE Compute GC graph $G$ and $\tilde{G}$ from origin and time reversed time series using Eq.\ref{eq15}.
   \STATE Get predict loss $\mathcal{L}_{{p(o)}}$, $\mathcal{L}_{{p(r)}}$, sparsity-inducing loss $\mathcal{L}_{{s(o)}}$, $\mathcal{L}_{{s(r)}}$ from origin and time reversed time series, respectively.   
   \IF{$\mathcal{L}_{{p(o)}}<\mathcal{L}_{{p(r)}}$ \textbf{AND} {$\mathcal{L}_{{s(o)}}<\mathcal{L}_{{s(r)}}$}} 
   \STATE $\hat{G}=G$
   \ELSIF{$\mathcal{L}_{{p(o)}}>\mathcal{L}_{{p(r)}}$ \textbf{AND} {$\mathcal{L}_{{s(o)}}>\mathcal{L}_{{s(r)}}$}}
   \STATE $\hat{G}=\tilde{G}$
   \ELSE
   \FOR{$i=1$ {\bfseries to} $p$}
   \FOR{$j=1$ {\bfseries to} $p$}
   \IF{$abs(G_{i,j}-\tilde{G}_{i,j})<\theta$}
   \STATE $\hat{G}_{i,j}=\frac{1}{2}(G_{i,j}+\tilde{G}_{i,j})$
   \ELSE 
   \STATE $\hat{G}_{i,j}=max(G_{i,j},\tilde{G}_{i,j})$
   \ENDIF
   \ENDFOR
   \ENDFOR
   \ENDIF
   \STATE {\bfseries return} $\bf \hat{G}$.
\end{algorithmic}
\end{algorithm}

\section{Experiment}
\label{experiment}
In this section, we present the performance of KANGCI on four widely used benchmark datasets: Lorenz-96, Gene regulatory networks, fMRI BOLD signals, and VAR. Comparative experiments are conducted against several state-of-the-art models, including cMLP \& cLSTM \cite{1tank2021neural}, TCDF \cite{2nauta2019causal}, eSRU \cite{3khanna2019economy}, GVAR \cite{4marcinkevivcs2021interpretable}, NAVAR (MLP) \& NAVAR (LSTM) \cite{8bussmann2021neural}, CUTS+ \cite{5cheng2024cuts+}, JGC \cite{22suryadi2023granger}, and JRNGC \cite{6zhoujacobian}. Moreover, we conduct additional experiments on real-world EEG signals to validate the effectiveness of KANGCI in practical applications. The corresponding results are provided in Section \ref{eeg}.

In alignment with prior studies, the model performances are evaluated using the area under the receiver operating characteristic curve (AUROC). Notably, in the evaluation of the Gene regulatory networks, only the off-diagonal elements of the Granger causality adjacency matrix are considered since the gold standard provided by the Gene regulatory networks does not account for self-causality. For the Lorenz-96, fMRI BOLD, and VAR datasets, all elements of the adjacency matrix are included.

\subsection{Lorenz-96}
Lorenz-96 is a mathematical model employed to investigate the dynamics of simplified atmospheric systems. Its behavior is governed by the following ordinary differential equation:  
\begin{equation}
\label{equation15}
\frac{\partial x_{t,i}}{\partial t} =-x_{t,i-1}\left( x_{t,i-2}-x_{t,i+1}\right)-x_{t,i}+F
\end{equation}
where $F$ represents the external forcing term in the system, and $p$ denotes the spatial dimension of the system. The increase in $F$ results in heightened system chaos, while the increase in $p$ enhances the spatial complexity of the system. We simulate $R=5$ replicates under the following three conditions : (1) $F=10$, $p=10$, $T=1000$ (low dimensionality, weak nonlinearity); (2) $F=40$, $p=40$, $T=1000$ (high dimensionality, strong nonlinearity); (3) $F=40$, $p=40$, $T=500$ (limited observations).

\begin{table}[!htbp]
  \centering
  \caption{AUROC of the Lorenz-96 dataset.}
    \resizebox{0.48\textwidth}{!}{
    \begin{tabular}{cccc}
    \toprule
    \toprule
    \multirow{3}{*}{Models} &   \multicolumn{3}{c}{AUROC}  \\
     \cmidrule{2-4}
    & \makecell{$p=10$, $F=10$\\$T=1000$} & \makecell{$p=40$, $F=40$\\$T=1000$} & \makecell{$p=40$, $F=40$\\$T=500$}  \\
    \midrule
    \midrule
        cMLP            & \numstd{0.983}{0.003}     & \numstd{0.867}{0.025}             & \numstd{0.843}{0.036}   \\
        cLSTM           & \numstd{0.978}{0.004}     & \numstd{0.943}{0.027}             & \numstd{0.863}{0.044}   \\
        TCDF            & \numstd{0.879}{0.011}     & \numstd{0.674}{0.039}             & \numstd{0.565}{0.041}   \\
        eSRU            & \bf \numstd{1.0}{0.00}    & \numstd{0.973}{0.012}             & \numstd{0.953}{0.025}   \\
        GVAR            & \bf \numstd{1.0}{0.00}    & \numstd{0.951}{0.016}             & \numstd{0.941}{0.022}   \\
        NAVAR (MLP)     & \underline{\numstd{0.993}{0.004}}     & \numstd{0.843}{0.033} & \numstd{0.787}{0.054}   \\
        NAVAR (LSTM)    & \numstd{0.993}{0.006}     & \numstd{0.821}{0.045}             & \numstd{0.791}{0.056}   \\
        JGC             & \numstd{0.994}{0.005}     & \numstd{0.944}{0.037}             & \numstd{0.927}{0.053}   \\
        CUTS+           & \bf \numstd{1.0}{0.00}    & \underline{\numstd{0.989}{0.003}} & \underline{\numstd{0.961}{0.012}}\\
        JRNGC           & \bf \numstd{1.0}{0.00}    & \numstd{0.979}{0.012}             & \numstd{0.956}{0.023}    \\
        \textbf{KANGCI} & \bf \numstd{1.0}{0.00}    & \bf \numstd{0.995}{0.002}         & \bf \numstd{0.972}{0.014}   \\
    \bottomrule
    \bottomrule
    \end{tabular}}%
  \label{tab-lorenz}%
\end{table}%
Table \ref{tab-lorenz} presents the Granger causality inference performance of each model under three conditions. For the scenario where $p=10$, $F=10$, and $T=1000$, all methods, except for TCDF, effectively infer the causal relationships. KANGCI, eSRU, GVAR, CUTS+, and JRNGC achieve an AUROC of 1.0. However, when  $p=40$, $F=40$, causal inference becomes more challenging, particularly as the time series length decreases. Under these conditions, the performance of cMLP, cLSTM, and NAVAR declines significantly. KANGCI achieves the highest AUROC (0.995 and 0.972, respectively). In summary, KANGCI exhibits superior performance on the Lorenz-96 dataset.

\subsection{Gene regulatory networks}
\subsubsection{Dream-3}
The second dataset is the DREAM-3 in Silico Network Challenge, available at \url{https://gnw.sourceforge.net/dreamchallenge.html}. The dataset provides complex and nonlinear time series for evaluating the performance of Granger causality models. It consists of five sub-datasets: two corresponding to E.coli (\textit{E.coli-1}, \textit{E.coli-2}) and three to Yeast (\textit{Yeast-1}, \textit{Yeast-2}, \textit{Yeast-3}). Each sub-dataset has a distinct ground-truth Granger causality network and includes $p=100$ time series, representing the expression levels of $n=100$ genes. Each time series comprises 46 replicates, sampled at 21 time points, yielding a total of 966 observations.

\begin{table}[!htbp]
  \centering
  \caption{AUROC of the Dream-3 dataset, T=966, p=100}
    \resizebox{0.48\textwidth}{!}{
    \begin{tabular}{cccccc}
    \toprule
    \toprule
    \multirow{2}{*}{Models} &   \multicolumn{5}{c}{AUROC}  \\
     \cmidrule{2-6}
     & Ecoli-1 & Ecoli-2 & Yeast-1 & Yeast-2 & Yeast-3  \\
    \midrule
    \midrule
        cMLP            & 0.648 & 0.568 & 0.585 & 0.511 & 0.531    \\
        cLSTM           & 0.651 & 0.609 & 0.579 & 0.524 & 0.552    \\
        TCDF            & 0.615 & 0.621 & 0.581 & 0.567 & 0.565   \\
        eSRU            & 0.660 & 0.636 & 0.631 & 0.561 & 0.559   \\
        GVAR            & 0.652 & 0.634 & 0.623 & 0.57 & 0.554   \\
        NAVAR (MLP)     & 0.557 & 0.577 & 0.652 & 0.573 & 0.548   \\
        NAVAR (LSTM)    & 0.544 & 0.473 & 0.497 & 0.477 & 0.466  \\
        JGC             & 0.522 & 0.536 & 0.611 & 0.558 & 0.531  \\
        CUTS+           & \underline{0.703} & 0.675 & \underline{0.661} & \textbf{0.612} & 0.554   \\
        JRNGC           & 0.666 & \underline{0.678} & 0.650 & \underline{0.597} & \underline{0.560}   \\
        \textbf{KANGCI}  & \textbf{0.758}  & \textbf{0.680} & \textbf{0.667}  & 0.552 & \textbf{0.562}   \\
    \bottomrule
    \bottomrule
    \end{tabular}}%
  \label{tab-dream3}%
\end{table}%

The results of the Dream-3 dataset are shown in Table~\ref{tab-dream3}. The performance of all models drops significantly compared to the Lorenz-96 dataset since the Dream-3 dataset contains 100 channels and carries additional noise, which leads to frequent overfitting of the models. Our model emerges as the top-performance model among its counterparts in four out of five sub-datasets. Specifically, the AUROC of the KANGCI in \textit{E.coli-1}, \textit{E.coli-2}, \textit{Yeast-1}, and \textit{Yeast-3} are 0.758, 0.680, 0.667 and 0.562, respectively. This further proves the effectiveness of our method in identifying sparse Granger causality in high-dimensional, noisy time series.

\subsubsection{Dream-4}
The third dataset is the DREAM-4 in silico challenge. Analogous to the DREAM-3 dataset, it consists of five sub-datasets, each containing $p=100$ time series. However, each time series in DREAM-4 only includes 10 replicates sampled at 21 time points, yielding a total of 210 observations. This is substantially fewer than the 966 observations provided by the DREAM-3 dataset. Therefore, Dream-4 dataset challenges the inference performance of each model in scenarios with a limited number of time series observations.

\begin{table}[!htbp]
  \centering
  \caption{AUROC of the Dream-4 dataset, T=210, p=100}
    \resizebox{0.48\textwidth}{!}{
    \begin{tabular}{cccccc}
    \toprule
    \toprule
    \multirow{2}{*}{Models} &   \multicolumn{5}{c}{AUROC}  \\
     \cmidrule{2-6}
    & Gene-1 & Gene-2 & Gene-3 & Gene-4 & Gene-5  \\    
    \midrule
    \midrule
        cMLP            & 0.652 & 0.522 & 0.509 & 0.511 & 0.531    \\
        cLSTM           & 0.633 & 0.509 & 0.498 & 0.524 & 0.552    \\
        TCDF            & 0.598 & 0.491 & 0.467 & 0.567 & 0.565   \\
        eSRU            & 0.647 & 0.554 & 0.545 & 0.561 & 0.559   \\
        GVAR            & 0.662 & 0.569 & 0.565 & 0.578 & 0.554   \\
        NAVAR (MLP)     & 0.591 & 0.522 & 0.507 & 0.543 & 0.548   \\
        NAVAR (LSTM)    & 0.587 & 0.514 & 0.525 & 0.537 & 0.531  \\
        JGC             & 0.544 & 0.502 & 0.513 & 0.505 & 0.517  \\        
        CUTS+           & \underline{0.738} & \textbf{0.622}  & \underline{0.591} & 0.584  & \underline{0.594}  \\
        JRNGC           & 0.731 & \underline{0.613} & 0.583 & \underline{0.597} & 0.580   \\
        \textbf{KANGCI}  & \textbf{0.747}  & 0.591  & \textbf{0.602}  & \textbf{0.613} & \textbf{0.601}   \\
    \bottomrule
    \bottomrule
    \end{tabular}}%
  \label{tab-dream4}%
\end{table}%

Table~\ref{tab-dream4} shows the improved performance of KANGCI in inferring gene-gene interactions from limited time-series data, outperforming baseline models. Specifically, our model achieves the highest AUROCs in four of the five gene networks, with values of 0.747, 0.602, 0.613, and 0.601 for networks 1, 3, 4, and 5, respectively.  

\begin{table*}[h]
  \centering
  \caption{AUROC of the fMRI BOLD signals, Subject=50, T=50/100/200/2000/5000}
  \resizebox{1\textwidth}{!}{
    \begin{tabular}{ccccccccccccc}
    \toprule
    \toprule
    \multirow{2}{*}{Dateset} &   \multicolumn{10}{c}{AUROC}\\
    \cmidrule{2-12} 
     &  cMLP   &  cLSTM & TCDF  & eSRU  & GVAR & NAVAR (MLP)  & NAVAR (LSTM) & JGC & CUTS+ & JRNGC & \textbf{KANGCI} \\
    \midrule
    \midrule
    Sim1  &\numstd{0.746}{0.04} &\numstd{0.689}{0.05} &\numstd{0.806}{0.03}  &\numstd{0.729}{0.04}  &\numstd{0.753}{0.05} &\numstd{0.723}{0.05} &\numstd{0.711}{0.05} &\numstd{0.812}{0.05} &\underline{\numstd{0.825}{0.04}} &\numstd{\textbf{0.829}}{0.04} &\numstd{0.815}{0.08}\\
    Sim2  &\numstd{0.733}{0.05} &\numstd{0.739}{0.04} &\numstd{0.823}{0.04}  &\numstd{0.756}{0.04}  &\numstd{0.723}{0.04} &\numstd{0.701}{0.03} &\numstd{0.694}{0.03} &\numstd{0.842}{0.02} &\underline{\numstd{0.851}{0.03}} &\numstd{0.833}{0.03} &\numstd{\textbf{0.857}}{0.03}\\
    Sim3  &\numstd{0.705}{0.06} &\numstd{0.735}{0.05} &\numstd{0.823}{0.03}  &\numstd{0.737}{0.04}  &\numstd{0.744}{0.05} &\numstd{0.703}{0.03} &\numstd{0.679}{0.04} &\underline{\numstd{0.866}{0.02}} &\numstd{0.859}{0.02} &\numstd{0.831}{0.03} &\numstd{\textbf{0.884}}{0.02}\\
    Sim4  &\numstd{0.685}{0.06} &\numstd{0.711}{0.05} &\numstd{0.814}{0.03}  &\numstd{0.722}{0.04}  &\numstd{0.738}{0.04} &\numstd{0.688}{0.04} &\numstd{0.647}{0.05} &\numstd{0.854}{0.02} &\numstd{0.869}{0.02} &\underline{\numstd{0.877}{0.01}} &\numstd{\textbf{0.916}}{0.01}\\
    Sim5  &\numstd{0.681}{0.05} &\numstd{0.691}{0.04} &\numstd{0.815}{0.03}  &\numstd{0.756}{0.04}  &\numstd{0.732}{0.03} &\numstd{0.794}{0.03} &\numstd{0.812}{0.04} &\numstd{0.838}{0.03} &\underline{\numstd{0.849}{0.04}} &\numstd{0.851}{0.05} &\numstd{\textbf{0.861}}{0.05}\\   
    Sim6  &\numstd{0.723}{0.15} &\numstd{0.738}{0.09} &\numstd{0.811}{0.02}  &\numstd{0.751}{0.03}  &\numstd{0.775}{0.03} &\numstd{0.826}{0.03} &\numstd{0.842}{0.03} &\numstd{0.881}{0.03} &\underline{\numstd{0.903}{0.03}} &\numstd{0.891}{0.03} &\numstd{\textbf{0.928}}{0.02}\\
    Sim7  &\numstd{0.708}{0.05} &\numstd{0.721}{0.04} &\numstd{0.809}{0.03}  &\numstd{0.781}{0.04}  &\numstd{0.744}{0.03} &\numstd{0.805}{0.03} &\numstd{0.827}{0.03} &\numstd{0.843}{0.03} &\underline{\numstd{0.866}{0.05}} &\numstd{0.841}{0.04} &\numstd{\textbf{0.902}}{0.04}\\
    Sim8  &\numstd{0.549}{0.15} &\numstd{0.522}{0.09} &\numstd{0.661}{0.08}  &\numstd{0.605}{0.09}  &\numstd{0.644}{0.07} &\numstd{0.601}{0.12} &\numstd{0.572}{0.11} &\numstd{0.629}{0.09} &\numstd{0.684}{0.08} &\underline{\numstd{0.712}{0.07}} &\numstd{\textbf{0.766}}{0.08}\\
    Sim9  &\numstd{0.667}{0.07} &\numstd{0.704}{0.09} &\numstd{0.789}{0.06}  &\numstd{0.710}{0.05}  &\numstd{0.679}{0.06} &\numstd{0.713}{0.08} &\numstd{0.727}{0.08} &\numstd{0.752}{0.07} &\underline{\numstd{0.819}{0.06}} &\numstd{0.806}{0.06} &\numstd{\textbf{0.830}}{0.08}\\
    Sim10 &\numstd{0.632}{0.07} &\numstd{0.648}{0.09} &\numstd{0.749}{0.06}  &\numstd{0.677}{0.11}  &\numstd{0.688}{0.08} &\numstd{0.709}{0.11} &\numstd{0.736}{0.12} &\numstd{0.675}{0.08} &\numstd{\textbf{0.799}}{0.07} &\numstd{0.774}{0.08} &\underline{\numstd{0.783}{0.07}}\\
    Sim11 &\numstd{0.726}{0.04} &\numstd{0.715}{0.03} &\numstd{0.785}{0.03}  &\numstd{0.737}{0.04}  &\numstd{0.742}{0.03} &\numstd{0.777}{0.03} &\numstd{0.784}{0.03} &\numstd{0.811}{0.03} &\numstd{0.816}{0.02} &\underline{\numstd{0.829}{0.03}} &\numstd{\textbf{0.837}}{0.03}\\
    Sim12 &\numstd{0.738}{0.05} &\numstd{0.751}{0.03} &\numstd{0.803}{0.04}  &\numstd{0.755}{0.03}  &\numstd{0.734}{0.04} &\numstd{0.796}{0.03} &\numstd{0.782}{0.03} &\numstd{0.802}{0.05} &\numstd{0.817}{0.04} &\underline{\numstd{0.832}{0.04}} &\numstd{\textbf{0.860}}{0.03}\\
    Sim13 &\numstd{0.596}{0.07} &\numstd{0.586}{0.04} &\numstd{0.714}{0.06}  &\numstd{0.655}{0.08}  &\numstd{0.676}{0.09} &\numstd{0.685}{0.08} &\numstd{0.693}{0.09} &\numstd{0.683}{0.09} &\numstd{0.716}{0.07} &\underline{\numstd{0.739}{0.07}} &\numstd{\textbf{0.757}}{0.08}\\
    Sim14 &\numstd{0.617}{0.08} &\numstd{0.654}{0.07} &\numstd{0.722}{0.06}  &\numstd{0.689}{0.07}  &\numstd{0.673}{0.09} &\numstd{0.716}{0.08} &\numstd{0.724}{0.07} &\numstd{0.741}{0.06} &\numstd{0.759}{0.07} &\underline{\numstd{0.761}{0.06}} &\numstd{\textbf{0.801}}{0.08}\\  
    Sim15 &\numstd{0.637}{0.10} &\numstd{0.647}{0.09} &\numstd{0.687}{0.06}  &\numstd{0.614}{0.09}  &\numstd{0.606}{0.08} &\numstd{0.664}{0.07} &\numstd{0.672}{0.09} &\numstd{0.692}{0.08} &\numstd{0.732}{0.08} &\numstd{\textbf{0.773}}{0.09} &\underline{\numstd{0.745}{0.08}}\\
    Sim16 &\numstd{0.604}{0.11} &\numstd{0.618}{0.13} &\numstd{0.706}{0.08}  &\numstd{0.653}{0.09}  &\numstd{0.635}{0.09} &\numstd{0.623}{0.07} &\numstd{0.646}{0.09} &\numstd{0.638}{0.12} &\underline{\numstd{0.729}{0.09}} &\numstd{0.713}{0.11} &\numstd{\textbf{0.758}}{0.09}\\
    Sim17 &\numstd{0.694}{0.05} &\numstd{0.686}{0.05} &\numstd{0.813}{0.03}  &\numstd{0.712}{0.04}  &\numstd{0.704}{0.05} &\numstd{0.769}{0.03} &\numstd{0.781}{0.04} &\numstd{0.794}{0.04} &\numstd{0.845}{0.03} &\underline{\numstd{0.862}{0.04}} &\numstd{\textbf{0.894}}{0.03}\\ 
    Sim18 &\numstd{0.657}{0.07} &\numstd{0.660}{0.07} &\numstd{0.778}{0.03}  &\numstd{0.684}{0.05}  &\numstd{0.691}{0.06} &\numstd{0.725}{0.06} &\numstd{0.748}{0.05} &\numstd{0.751}{0.06} &\underline{\numstd{0.831}{0.05}} &\numstd{\textbf{0.837}}{0.05} &\numstd{0.818}{0.06}\\ 
    Sim19 &\numstd{0.733}{0.05} &\numstd{0.772}{0.04} &\numstd{0.849}{0.03}  &\numstd{0.793}{0.05}  &\numstd{0.739}{0.06} &\numstd{0.779}{0.04} &\numstd{0.826}{0.04} &\numstd{0.847}{0.04} &\underline{\numstd{0.871}{0.03}} &\numstd{0.865}{0.03} &\numstd{\textbf{0.906}}{0.03}\\
    Sim20 &\numstd{0.750}{0.04} &\numstd{0.795}{0.09} &\numstd{0.861}{0.02}  &\numstd{0.822}{0.03}  &\numstd{0.765}{0.05} &\numstd{0.819}{0.03} &\numstd{0.853}{0.04} &\numstd{0.877}{0.02} &\underline{\numstd{0.915}{0.03}} &\numstd{0.898}{0.02} &\numstd{\textbf{0.921}}{0.03}\\   
    Sim21 &\numstd{0.651}{0.07} &\numstd{0.674}{0.08} &\numstd{0.753}{0.05}  &\numstd{0.707}{0.06}  &\numstd{0.719}{0.04} &\numstd{0.688}{0.05} &\numstd{0.702}{0.06} &\numstd{0.643}{0.08} &\underline{\numstd{0.786}{0.06}} &\numstd{0.767}{0.06} &\numstd{\textbf{0.812}}{0.07}\\
    Sim22 &\numstd{0.674}{0.06} &\numstd{0.682}{0.06} &\numstd{0.746}{0.05}  &\numstd{0.718}{0.07}  &\numstd{0.726}{0.05} &\numstd{0.649}{0.07} &\numstd{0.674}{0.06} &\numstd{0.661}{0.07} &\numstd{0.797}{0.05} &\underline{\numstd{0.801}{0.06}} &\numstd{\textbf{0.825}}{0.06}\\
    Sim23 &\numstd{0.574}{0.08} &\numstd{0.598}{0.09} &\numstd{0.662}{0.05}  &\numstd{0.619}{0.08}  &\numstd{0.624}{0.09} &\numstd{0.585}{0.09} &\numstd{0.592}{0.08} &\numstd{0.624}{0.09} &\numstd{0.641}{0.08} &\numstd{\textbf{0.705}}{0.09} &\underline{\numstd{0.671}{0.08}}\\
    Sim24 &\numstd{0.526}{0.09} &\numstd{0.547}{0.13} &\numstd{0.570}{0.06}  &\numstd{0.558}{0.06}  &\numstd{0.561}{0.08} &\numstd{0.529}{0.11} &\numstd{0.548}{0.12} &\numstd{0.534}{0.07} &\numstd{\textbf{0.611}}{0.07} &\numstd{0.581}{0.07} &\underline{\numstd{0.594}{0.09}}\\
    Sim25 &\numstd{0.627}{0.07} &\numstd{0.613}{0.05} &\numstd{0.681}{0.04}  &\numstd{0.633}{0.07}  &\numstd{0.641}{0.05} &\numstd{0.608}{0.06} &\numstd{0.595}{0.07} &\numstd{0.645}{0.04} &\numstd{0.707}{0.06} &\underline{\numstd{0.728}{0.06}} &\numstd{\textbf{0.763}}{0.08}\\
    Sim26 &\numstd{0.593}{0.07} &\numstd{0.588}{0.07} &\numstd{0.668}{0.07}  &\numstd{0.612}{0.06}  &\numstd{0.633}{0.07} &\numstd{0.590}{0.06} &\numstd{0.563}{0.06} &\numstd{0.634}{0.05} &\numstd{0.682}{0.08} &\underline{\numstd{0.701}{0.07}} &\numstd{\textbf{0.721}}{0.09}\\
    Sim27 &\numstd{0.642}{0.08} &\numstd{0.631}{0.06} &\numstd{0.699}{0.05}  &\numstd{0.644}{0.09}  &\numstd{0.695}{0.06} &\numstd{0.626}{0.07} &\numstd{0.598}{0.09} &\numstd{0.656}{0.08} &\numstd{0.708}{0.07} &\underline{\numstd{0.727}{0.06}} &\numstd{\textbf{0.753}}{0.08}\\
    Sim28 &\numstd{0.688}{0.06} &\numstd{0.658}{0.05} &\numstd{0.762}{0.04}  &\numstd{0.709}{0.06}  &\numstd{0.735}{0.05} &\numstd{0.641}{0.04} &\numstd{0.603}{0.04} &\numstd{0.743}{0.07} &\numstd{0.764}{0.08} &\underline{\numstd{0.772}{0.06}} &\numstd{\textbf{0.821}}{0.07}\\
    \bottomrule
    \bottomrule
    \end{tabular}}%
  \label{tab-fmribold}%
\end{table*}%

\subsection{fMRI BOLD signals}
The fourth dataset is the simulated fMRI BOLD signals generated using the dynamic causal model (DCM) with the nonlinear balloon model for vascular dynamics. Each data includes multiple time series corresponding to different brain regions of interest (ROIs). Notably, the fMRI BOLD dataset contains 28 sub-datasets, each comprising 50 subjects and including distinct features. However, previous studies have typically utilized few subjects from few simulations (e.g., sim-3, sim-4) for model evaluation, which is inadequate for comprehensively assessing model performance on the fMRI dataset. In this study, we address this limitation by conducting a thorough evaluation using all subjects from all simulations (a total of 1,400 subjects). The dataset is shared at \url{https://www.fmrib.ox.ac.uk/datasets/netsim/index.html}. Table~\ref{tab-fmribold} presents the comparison results of all simulations.

Comparative experiments conducted on the fMRI BOLD dataset demonstrate that only TCDF, JGC, JRNGC, CUTS+, and KANGCI effectively infer Granger causality across all simulations and subjects.  Among these methods, KANGCI achieved superior performance in 22 out of 28 simulations, covering various complex scenarios such as global mean confusion, mixed time series, shared inputs, backward connections, cyclic connections, and time lags. In contrast, JRNGC and CUTS+ exhibited better performance in simulations with varying connection strengths (e.g., sim 15, 22, 23). Furthermore, given the inclusion of noise and randomness (with a standard deviation of 0.5 seconds in the hemodynamic response function delay) and the limited sampling points ($T=200$) in most cases, the proposed model can more effectively infer Granger causality under noisy and data-constrained conditions compared to existing baseline models.

\subsection{VAR}
The fifth dataset is the VAR model. For a $p$-dimensional time series $x_{t}$, the VAR model is given by:
\begin{equation}
x_{t}=A^{\left( 1\right) }x_{t-1}+A^{(2)}x_{t-2}+, \ldots , +A^{\left( k\right) }x_{t-k}+u_{t}
\end{equation}
where $(A^{(1)},A^{(2)},\ldots,A^{(k})$ are regression coefficients matrices and $u_{t}$ is a vector of errors with Gaussian distribution. We define $sparsity$ as the percentage of non-zero coefficients in $A^{(i)}$, and different $sparsity$ represent different quantities of Granger causality interaction in the VAR model. The comparison results of the VAR dataset are presented in Table \ref{tab-var}.

\begin{table}[!htbp]
  \centering
  \caption{AUROC of the VAR dataset.}
    \resizebox{0.48\textwidth}{!}{
    \begin{tabular}{cccc}
    \toprule
    \toprule
    \multirow{4}{*}{Models} &   \multicolumn{3}{c}{AUROC}  \\
     \cmidrule{2-4}
    & \makecell{$p=10, T=1000$\\$sparsity=0.2$\\$lag=3$} & \makecell{$p=10, T=1000$\\$sparsity=0.3$\\$lag=3$} & \makecell{$p=10, T=1000$\\$sparsity=0.2$\\$lag=5$}\\
    \midrule
    \midrule
        cMLP            & \bf  \numstd{1}{0.00}                 & \numstd{0.947}{0.004}             & \numstd{0.986}{0.002}\\
        cLSTM           & \numstd{0.986}{0.004}                 & \numstd{0.921}{0.004}             & \numstd{0.961}{0.003}\\
        TCDF            & \numstd{0.879}{0.011}                 & \numstd{0.759}{0.007}             & \numstd{0.823}{0.006}\\
        eSRU            & \bf \numstd{1.0}{0.00}                & \numstd{0.995}{0.001}             & \bf  \numstd{1.0}{0.00}\\
        GVAR            & \bf \numstd{1.0}{0.00}                & \numstd{0.992}{0.002}             & \bf  \numstd{1.0}{0.00}\\
        NAVAR (MLP)     & \underline{\numstd{0.993}{0.002}}     & \numstd{0.986}{0.003}             &  \underline{\numstd{0.992}{0.002}}\\
        NAVAR (LSTM)    & \underline{\numstd{0.993}{0.002}}     & \numstd{0.963}{0.004}             & \numstd{0.987}{0.002}\\
        JGC             & \bf \numstd{1.0}{0.00}                & \numstd{0.995}{0.002}             & \bf \numstd{1.0}{0.00} \\
        CUTS+           & \bf \numstd{1.0}{0.00}                & \bf \numstd{1.0}{0.00}            & \bf \numstd{1.0}{0.00} \\
        JRNGC           & \bf \numstd{1.0}{0.00}                & \underline{\numstd{0.997}{0.001}} & \bf \numstd{1.0}{0.00}    \\
        \textbf{KANGCI} & \bf \numstd{1.0}{0.00}               & \numstd{0.993}{0.003}              & \bf \numstd{1.0}{0.00}    \\
    \bottomrule
    \bottomrule
    \end{tabular}}%
  \label{tab-var}%
\end{table}%

The comparison results reveal that all models, with the exception of TCDF, effectively infer Granger causality from the VAR dataset. Among these, CUTS+ demonstrates the highest performance, achieving an AUROC of 1.0 in three scenarios. KANGCI, JRNGC, JGC, GVAR, and e-SRU achieve an AUROC of 1.0 in two scenarios. For cMLP and cLSTM, the performance decreases slightly when lag or sparsity are varied.

\section{Experiment on real-world EEG signals}
\label{eeg}

The experiment in Section \ref{experiment} shows that the proposed model can effectively infer Granger causality from time series. However, these experiments are conducted on simulated datasets, and the applicability and effectiveness of the model on real-world data still need to be further validated. Therefore, in this section, we aim to verify the effectiveness of KANGCI on real-world EEG signals.

\subsection{Data collection and preprocessing}
\begin{figure}[h]
\centering
\includegraphics[width=1\columnwidth]{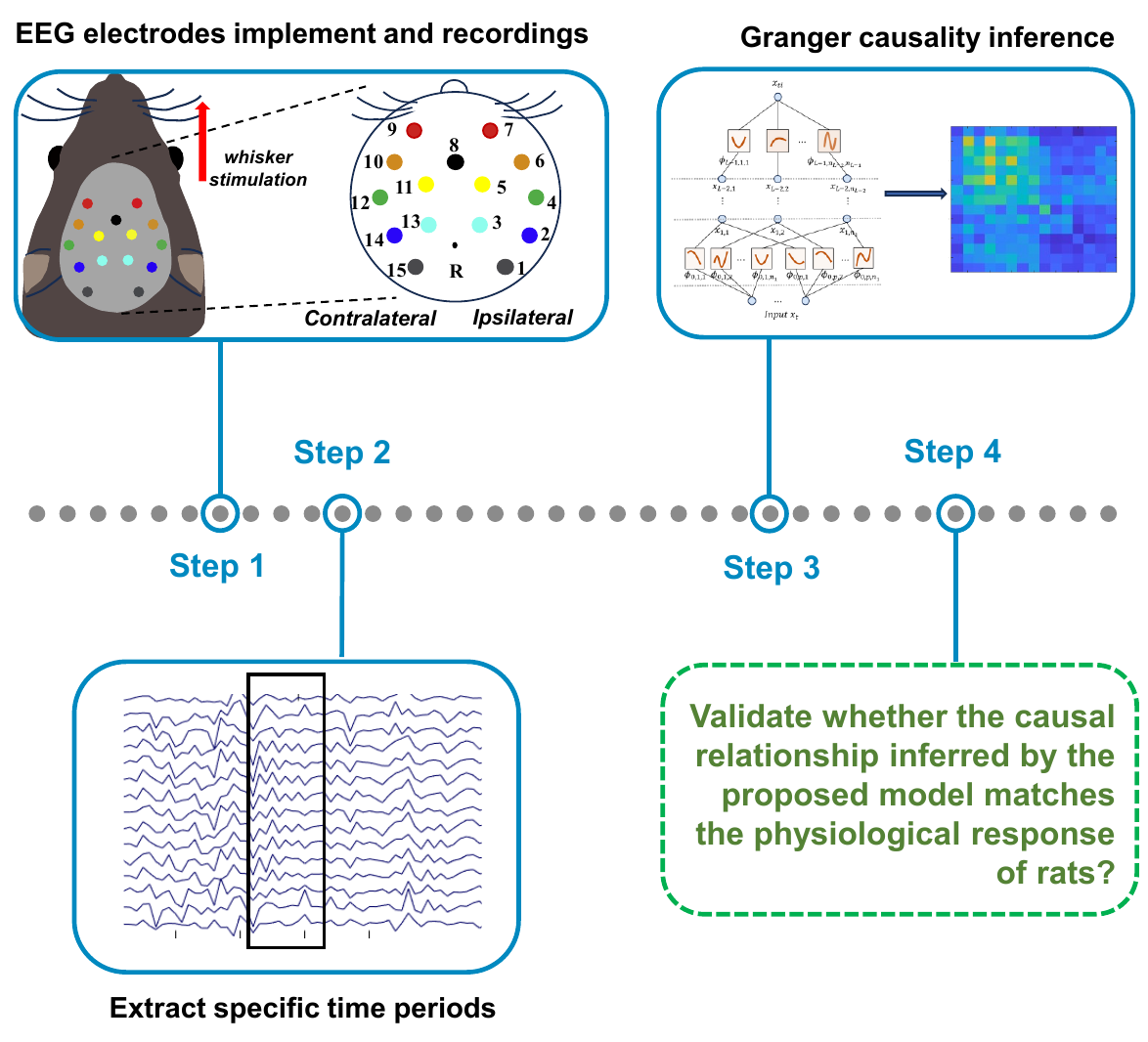}
\caption{The analysis pipeline of real-world whisker stimulation rat EEG signals. Step 1: EEG electrode positions. A solenoid is used to stimulate the unilateral whisker of the rat, with nodes 1-7 representing the ipsilateral electrodes to stimulation and nodes 9-15 representing the contralateral electrodes to stimulation. Step 2: Extracting the time period to be analyzed. Step 3: Inferring the Granger causality from time series using KANGCI. Step 4: Validating whether the inferred Granger causality matches the physiological response of rats.}
\label{fig2}
\end{figure}

We utilize the EEG dataset provided by \citet{32pagnotta2018assessing}, which comprises somatosensory evoked potentials (SEPs) induced by whisker stimulation of 10 Wistar rats. These rats are anesthetized and subjected to unilateral whisker stimulation via a solenoid for 500 ms across 100 trials. SEPs are recorded using a stainless steel electrode grid positioned on the skulls of the rats. SEP signals are sampled at 2000 Hz using a bandpass filtered between 1-500 Hz. The signals contain a time of -100 ms pre- to 200 ms post-stimulation. The analysis pipeline is illustrated in Fig.\ref{fig2}.

For the EEG prepossessing, we apply two criteria to identify and exclude trials potentially affected by artifacts. Specifically, a trial is considered contaminated if it meets one of two conditions: (1) the signal variance is higher in the pre-stimulation than in the post-stimulation for over three channels; (2) the signal during the pre-stimulation period exceeds a threshold of 200 $\mu$V in at least one channel \cite{33plomp2014physiological,34trongnetrpunya2016assessing,35barnett2011behaviour}. Furthermore, we do not apply any additional filter, as prior research has indicated that filters could compromise the integrity of the informational content and order of data, subsequently affecting the inference of Granger causality \cite{36pullon2020granger}.

\subsection{Evaluation criteria}

We evaluate the performance of KANGCI based on three previously proposed criteria (criteria 2-4) \cite{33plomp2014physiological,37pagnotta2018benchmarking} and two additional criteria (criteria 1, 5), which collectively examine five distinct characteristics anticipated in the cortical network comprising 15 nodes.

1. \textbf{Information flow loss induced by anesthesia}: Information flow between brain regions is essential for sustaining awake consciousness, and anesthesia would induce the loss of information flow (causal relationship), leading to loss of consciousness \cite{36pullon2020granger}. Consequently, the first criterion is to assess whether the model can detect the absence of Granger causality during the -100 to 0 ms epoch of anesthesia.

2. \textbf{Latency differences in sensory cortices}: Stimulation on rat whiskers would activate the primary sensory cortex (S1). However, the latencies of the ipsilateral S1 (iS1, node 4) and contralateral S1 (cS1, node 12) are different (cS1 is about 14ms, iS1 is about 26ms). Therefore, whether the model can infer Granger causality originated from the cS1 and iS1 regions during 10-20ms and 20-30ms, respectively, is the gold standard for evaluating the effectiveness of the model.

3. \textbf{Causal driving identification}: Does the model identify the cS1 and iS1 as the main causal driving in the corresponding time period?

4. \textbf{Causality from cS1 to contralateral Regions}: Does the model accurately identify the Granger causality from cS1 to the contralateral frontal (node 10) and parietal (node 14) regions?

5. \textbf{Causality from iS1 to ipsilateral Regions}: Does the model accurately identify the Granger causality from iS1 to the ipsilateral frontal (node 2) and parietal (node 6) regions?

\subsection{Results}

\begin{figure}[h]
\centering
\includegraphics[width=0.95\columnwidth]{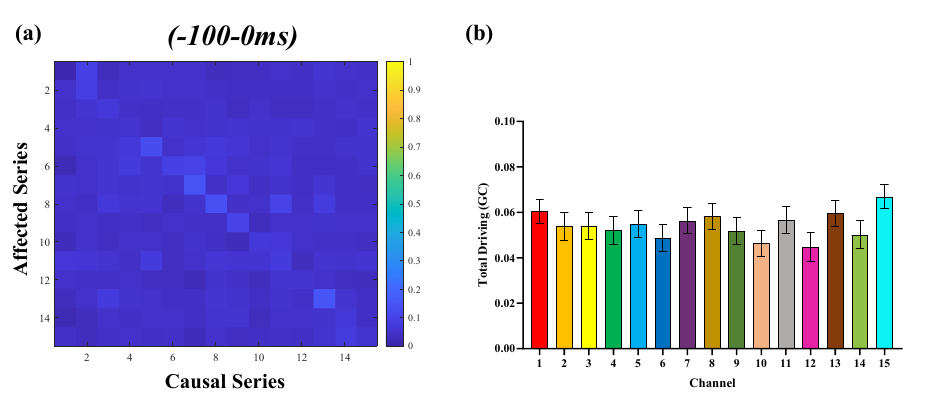}
\caption{(a) The inferred Granger causality in epoch -100-0 ms. (b) The Granger causality driving of each channel.}
\label{fig3}
\end{figure}

As shown in Fig.\ref{fig3}(a), KANGCI does not detect any significant causal relationship from -100 to 0 milliseconds pre-stimulation, and the causal driving of each channel is only around 0.06 (Fig.\ref{fig3}(b)). These results indicate that the model detects the absence (loss) of Granger causality caused by anesthesia during the non-stimulation period (criterion 1).

\begin{figure}[h]
\centering
\includegraphics[width=0.95\columnwidth]{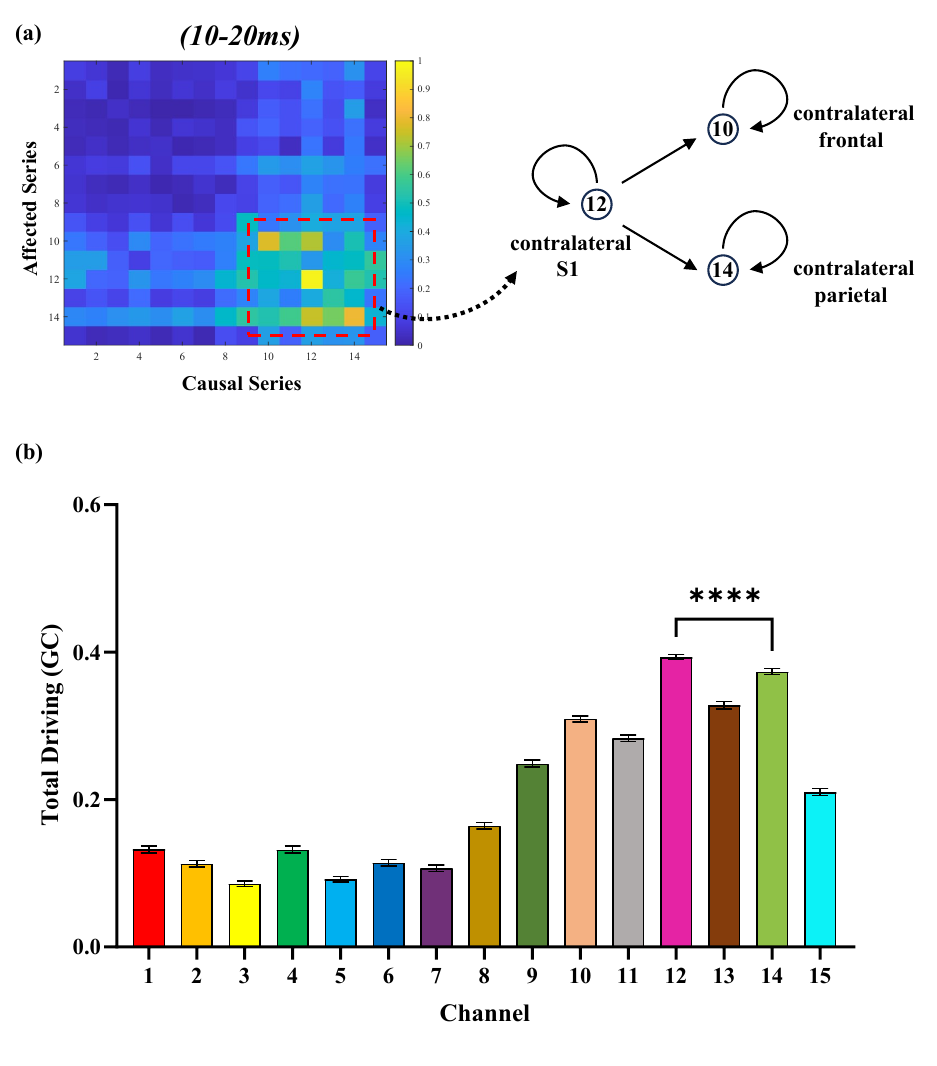}
\caption{(a) The inferred Granger causality in epoch 10-20 ms. (b) The causal driving of each channel.}
\label{fig4}
\end{figure}

\begin{figure}[h]
\centering
\includegraphics[width=0.95\columnwidth]{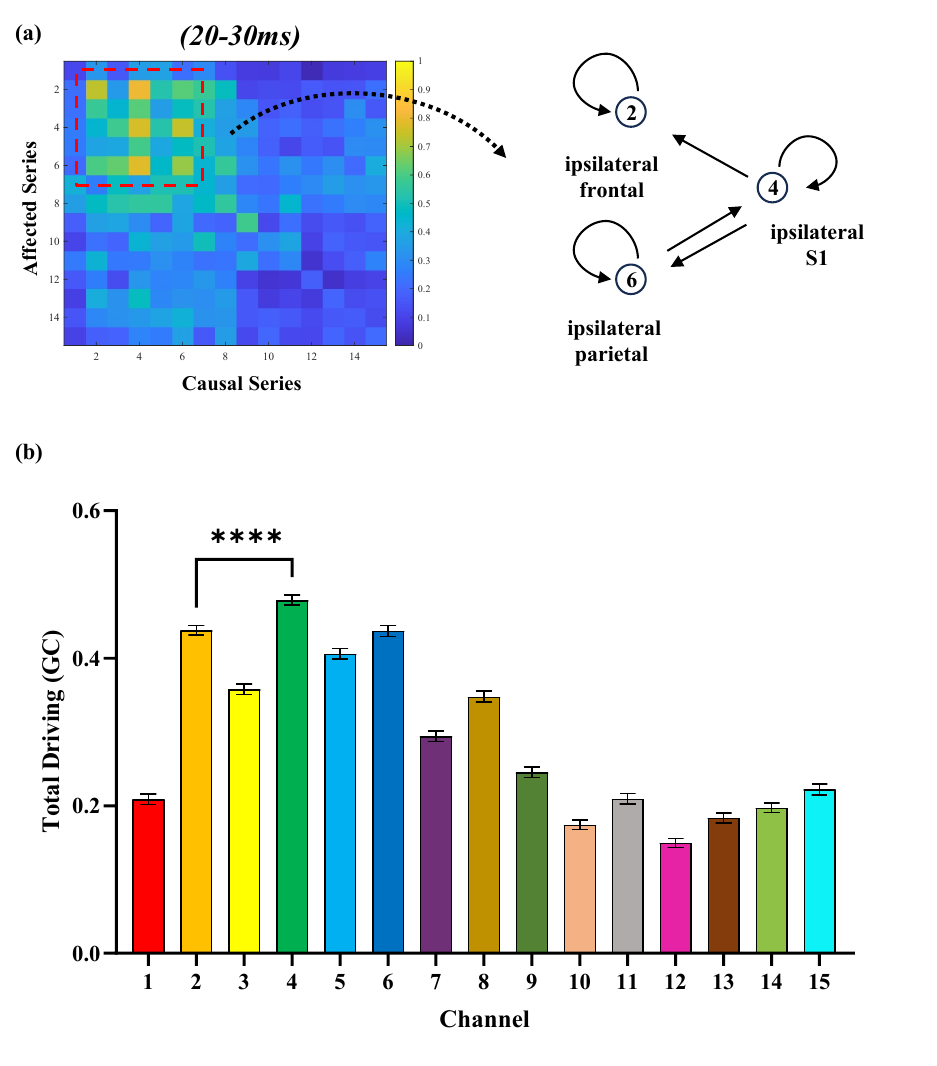}
\caption{(a) The inferred Granger causality in epoch 20-30 ms. (b) The causal driving of each channel.}
\label{fig5}
\end{figure}

Furthermore, KANGCI effectively identifies the causal relationship from cS1 to the contralateral frontal and parietal regions during the 10-20 ms epoch, as illustrated in Fig.\ref{fig4}(a). Meanwhile, we conduct statistical analysis of the causal driving for all channels using one-way ANOVA (Fig.\ref{fig4}(b)). The result shows that the causal driving of cS1 is significantly greater than that of all other nodes (p$<$0.0001), indicating that cS1 is the primary causal driver during 10-20 ms. These findings satisfy evaluation criteria 2, 3, and 4.

During the time period of 20-30 ms, KANGCI successfully identifies the causal relationship from iS1 to the ipsilateral frontal and parietal regions, as depicted in Fig.\ref{fig5}(a). One-way ANOVA also reveals that the causal driving of iS1 is significantly larger than that of all other nodes (p$<$0.0001)  (Fig.\ref{fig5}(b)), indicating that iS1 is the primary causal driving during 20-30 ms. Consequently, these results match evaluation criteria 2, 3, and 5.

Therefore, these findings collectively highlight KANGCI's efficiency in identifying distinct causal relationships across various time periods, validating KANGCI's ability to infer Granger causality from real-world EEG signals.

\section{Conclusion}

In this study, we propose a novel neural network-based Granger causality model, termed Granger Causality inference Kolmogorov-Arnold Networks (KANGCI). The model leverages the base weights of KAN layers, incorporating sparsity-inducing penalty and ridge regularization to infer the causal relationship. In addition, we develop an algorithm grounded in time-reverse Granger causality to mitigate spurious connections and enhance inference performances. Extensive experiments on Lorenz-96, Gene regulatory networks, fMRI BOLD, VAR, and real-world EEG signals validate that KANGCI can effectively infer Granger causality relationships from time series, outperforming the existing baselines. These results suggest that KANGCI brings a new avenue for Granger causality inference. We anticipate that this model will inspire subsequent research to design more accurate and computationally efficient frameworks for causal inference.

\clearpage
\bibliography{example_paper}
\bibliographystyle{icml2025}

%%%%%%%%%%%%%%%%%%%%%%%%%%%%%%%%%%%%%%%%%%%%%%%%%%%%%%%%%%%%%%%%%%%%%%%%%%%%%%%
%%%%%%%%%%%%%%%%%%%%%%%%%%%%%%%%%%%%%%%%%%%%%%%%%%%%%%%%%%%%%%%%%%%%%%%%%%%%%%%
% APPENDIX
%%%%%%%%%%%%%%%%%%%%%%%%%%%%%%%%%%%%%%%%%%%%%%%%%%%%%%%%%%%%%%%%%%%%%%%%%%%%%%%
%%%%%%%%%%%%%%%%%%%%%%%%%%%%%%%%%%%%%%%%%%%%%%%%%%%%%%%%%%%%%%%%%%%%%%%%%%%%%%%
\clearpage
\appendix

%%%%%%%%%%%%%%%%%%%%%%%%%%%%%%%%%%%%%%%%%%%%%%%%%%%%%%%%%%%%%%%%%%%%%%%%%%%%%%%
%%%%%%%%%%%%%%%%%%%%%%%%%%%%%%%%%%%%%%%%%%%%%%%%%%%%%%%%%%%%%%%%%%%%%%%%%%%%%%%

\end{document}